\documentclass{article}
\usepackage{etex}
\usepackage{pgfplots}
\usetikzlibrary{plotmarks}
\definecolor{col1}{rgb}{1.00, 0.14, 0.14}
\definecolor{col2}{rgb}{0.13, 0.17, 0.53}
\definecolor{col3}{rgb}{0.99, 0.68, 0.38}
\definecolor{col4}{rgb}{0.67, 0.87, 0.64}
\usepackage{siunitx}

\usepackage{spconf,amsmath,graphicx}
\usepackage{enumitem}
\usepackage{amssymb}
\usepackage{booktabs}
\usepackage{multirow}
\usepackage{placeins}

\usepackage{pst-all}

\usepackage{hyperref}

\title{Simple Online and Realtime Tracking}
\name{Alex Bewley$^{\dagger}$\thanks{Thanks to ACARP for funding.}, Zongyuan Ge$^{\dagger}$, Lionel Ott$^{\diamond}$, Fabio Ramos$^{\diamond}$, Ben Upcroft$^{\dagger}$}
\address{Queensland University of Technology$^{\dagger}$, University of Sydney$^{\diamond}$}
\begin{document}
%
\maketitle

\begin{abstract}

This paper explores a pragmatic approach to multiple object tracking
where the main focus is to associate objects efficiently for online and
realtime applications. 
To this end, detection quality is identified as a key factor 
influencing tracking performance, where changing the detector can 
improve tracking by up to 18.9\%. 
Despite only using a rudimentary combination of
familiar techniques such as the Kalman Filter and Hungarian algorithm
for the tracking components, this approach achieves an accuracy
comparable to state-of-the-art online trackers. Furthermore, due to the
simplicity of our tracking method, the tracker updates at a rate of
\SI{260}{\Hz} which is over 20x faster than other state-of-the-art trackers. 

\end{abstract}

 
\section{Introduction}
\label{sec:intro}

This paper presents a lean implementation of a tracking-by-detection
framework for the problem of multiple object tracking (MOT) where
objects are detected each frame and represented as bounding boxes. 
In contrast to many batch based tracking approaches
\cite{Dicle2013,Rezatofighi2015,Kim2015}, this work is primarily
targeted towards online tracking where only detections from the previous
and the current frame are presented to the tracker. Additionally, a
strong emphasis is placed on efficiency for facilitating realtime
tracking and to promote greater uptake in applications such as 
pedestrian tracking for autonomous vehicles.

\begin{figure}[!tb]
	\centering
\begin{tikzpicture}

    \begin{semilogyaxis}[
        xlabel=Accuracy (MOTA),
        ylabel=Speed (Hz),
        title=Accuracy vs. Speed,
        log basis x=10,
        log ticks with fixed point,
        mark size=4,
        legend style={
            at={(0.5, -0.2)},
            anchor=north,
            legend columns=4,
            legend cell align=left,
            font=\small,
            draw=none,
        }
    ]

        \addplot [col1, only marks, mark=square*] coordinates {(32.4,   5.9)}; 
        \addplot [col1, only marks, mark=*] coordinates {(19.8, 112.1)}; 
        \addplot [col1, only marks, mark=triangle*] coordinates {(15.9,   0.7)}; 
        \addplot [col1, only marks, mark=diamond*] coordinates {(23.8,  32.6)}; 
        \addplot [col2, only marks, mark=square*] coordinates {(33.7,  11.5)}; 
        \addplot [col2, only marks, mark=*] coordinates {(14.0, 444.0)}; 
        \addplot [col2, only marks, mark=triangle*] coordinates {(18.2,   2.7)}; 
        \addplot [col2, only marks, mark=diamond*] coordinates {(33.4, 260.0)}node[black,above]{Proposed}; 
        \addplot [col3, only marks, mark=square*] coordinates {(30.3,   1.1)}; 
        \addplot [col3, only marks, mark=*] coordinates {(15.1,   1.7)}; 
        \addplot [col3, only marks, mark=triangle*] coordinates {(32.4,   0.7)}; 

        \addplot [black, dashed] coordinates {(13, 25) (36, 25)};

        \node [above] at (axis cs:15, 25){\small{$\uparrow$ Realtime}};

        \legend{TDAM, LP2D, TBD, JPDA, NOMT, DP\_NMS, SMOT, \textbf{SORT}, MDP, TC\_ODAL, MHT\_DAM};
    \end{semilogyaxis}

\end{tikzpicture}
	\caption
	{ Benchmark performance of the proposed method (\textbf{SORT}) in relation to several baseline trackers \cite{MOT2015}.
		Each marker indicates a tracker’s accuracy and speed measured in frames per second (FPS) [Hz], i.e. higher and more right is better. 
	}
	\label{fig:runtime_vs_mota}
\end{figure}
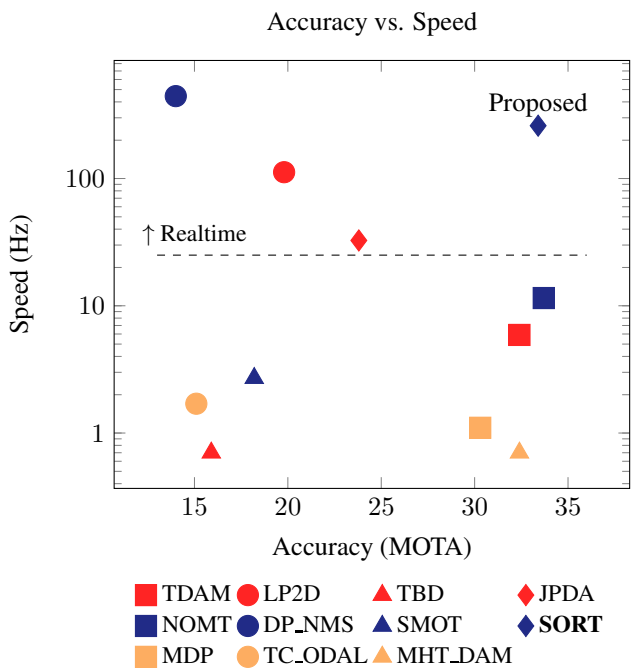

The MOT problem can be viewed as a data association problem where the 
aim is to associate detections across frames in a video sequence. 
To aid the data association process, trackers use various methods for 
modelling the motion \cite{Dicle2013, Yoon2015} and appearance 
\cite{Bewley2016_alextrac,Kim2015} of objects in the scene.
The methods employed by this paper were motivated through observations
made on a recently established visual MOT benchmark \cite{MOT2015}.
Firstly, there is a resurgence of mature data association techniques
including Multiple Hypothesis Tracking (MHT) \cite{Reid_1979,Kim2015}
and Joint Probabilistic Data Association (JPDA) \cite{Rezatofighi2015} 
which occupy many of the top positions of the MOT benchmark. Secondly, the
only tracker that does not use the Aggregate Channel Filter (ACF)
\cite{Dollar2014PAMI} detector is also the top ranked tracker,
suggesting
that detection quality could be holding back the other trackers.
Furthermore, the trade-off between accuracy and speed appears quite 
pronounced, since the speed of most accurate trackers is considered too
slow for realtime applications (see Fig. \ref{fig:runtime_vs_mota}).
With the prominence of traditional data association techniques among the
top online and batch trackers along with the use of different detections
used by the top tracker, this work explores how simple MOT can be and
how well it can perform.

Keeping in line with Occam's Razor, appearance features beyond the
detection component are ignored in tracking and only the bounding box
position and size are used for both motion estimation and data
association. Furthermore, issues regarding short-term and long-term
occlusion are also ignored, as they occur very rarely and their explicit
treatment introduces undesirable complexity into the tracking framework.
We argue that incorporating complexity in the form of object
re-identification adds significant overhead into the tracking framework
-- potentially limiting its use in realtime applications.

This design philosophy is in contrast to many proposed visual trackers
that incorporate a myriad of components to handle various edge cases and
detection errors \cite{Oh2004,Perera2006,Choi2015,Xiang2015}. This work
instead focuses on efficient and reliable handling of the common
frame-to-frame associations. Rather than aiming to be robust to
detection errors, we instead exploit recent advances in visual object
detection to solve the detection problem directly. This is demonstrated
by comparing the common ACF pedestrian detector \cite{Dollar2014PAMI}
with a recent convolutional neural network (CNN) based detector
\cite{Ren2015}. Additionally, two classical yet extremely efficient
methods, Kalman filter \cite{Kalman1960} and Hungarian method
\cite{Hungarian}, are employed to handle the motion prediction and data
association components of the tracking problem respectively. This
minimalistic formulation of tracking facilitates both efficiency and
reliability for online tracking, see Fig. \ref{fig:runtime_vs_mota}. In
this paper, this approach is only applied to tracking pedestrians in
various environments, however due to the flexibility of CNN based
detectors~\cite{Ren2015}, it naturally can be generalized to other
objects classes.

The main contributions of this paper are:
\begin{itemize}[noitemsep,nolistsep]
    \item We leverage the power of CNN based detection in the context of
        MOT.
    \item A pragmatic tracking approach based on the Kalman filter and
        the Hungarian algorithm is presented and evaluated on a recent
        MOT benchmark. 
    \item Code will be open sourced to help establish a baseline method
        for research experimentation and uptake in collision avoidance
        applications.
\end{itemize}

This paper is organised as follows: Section \ref{sec:literature}
provides a short review of related literature in the area of multiple
object tracking. Section \ref{sec:method} describes the proposed lean
tracking framework before the effectiveness of the proposed framework on
standard benchmark sequences is demonstrated in Section
\ref{sec:experiments}. Finally, Section \ref{sec:conclusion} provides a
summary of the learnt outcomes and discusses future improvements.

\section{Literature Review}
\label{sec:literature}

Traditionally MOT has been solved using Multiple Hypothesis Tracking
(MHT) \cite{Reid_1979} or the Joint Probabilistic Data
Association (JPDA) filters \cite{Bar-Shalom1987,Rezatofighi2015}, which
delay making difficult decisions while there is high uncertainty over
the object assignments. The combinatorial complexity of these
approaches is exponential in the number of tracked objects making them
impractical for realtime applications in highly dynamic environments.
Recently, Rezatofighi et al. \cite{Rezatofighi2015}, revisited the JPDA
formulation \cite{Bar-Shalom1987} in visual MOT with the goal to address
the combinatorial complexity issue with an efficient approximation of
the JPDA by exploiting recent developments in solving integer programs.
Similarly, Kim et al. \cite{Kim2015} used an appearance model for each
target to prune the MHT graph to achieve state-of-the-art performance.
However, these methods still delay the decision making which makes them
unsuitable for online tracking.

Many online tracking methods aim to build appearance models of either the 
individual objects themselves \cite{Bae2014,Min2015,Xiang2015} or a global 
model \cite{Bewley2014, Choi2015,Yoon2015,Bewley2016_alextrac} through 
online learning.
In addition to appearance models, motion is often incorporated to assist 
associating detections to tracklets \cite{Dicle2013, Bewley2014,Yoon2015, Choi2015}.
When considering only one-to-one correspondences modelled as bipartite 
graph matching, globally optimal solutions such as the Hungarian algorithm 
\cite{Hungarian} can be used \cite{Perera2006,Geiger2014PAMI}. 


The method by Geiger et al. \cite{Geiger2014PAMI} uses the Hungarian algorithm
\cite{Hungarian} in a two stage process. First, tracklets are formed by
associating detections across adjacent frames where both geometry and
appearance cues are combined to form the affinity matrix. Then, the
tracklets are associated to each other to bridge broken trajectories
caused by occlusion, again using both geometry and appearance cues. This
two step association method restricts this approach to batch
computation. Our approach is inspired by the tracking component of 
\cite{Geiger2014PAMI}, however we simplify the association to a single 
stage with basic cues as described in the next section. 


\section{Methodology}
\label{sec:method}

The proposed method is described by the key components of detection,
propagating object states into future frames, associating current
detections with existing objects, and managing the lifespan of tracked
objects. 


\subsection{Detection}
To capitalise on the rapid advancement of CNN based detection, we utilise the Faster Region CNN (\textbf{FrRCNN}) detection framework \cite{Ren2015}. 
\textbf{FrRCNN} is an end-to-end framework that consists of two stages.
The first stage extracts features and proposes regions for the second stage which then classifies the object in the proposed region.
The advantage of this framework is that parameters are shared between
the two stages creating an efficient framework for detection.
Additionally, the network architecture itself can be swapped to any
design which enables rapid experimentation of different architectures to
improve the detection performance.


Here we compare two network architectures provided with \textbf{FrRCNN},
namely the architecture of Zeiler and Fergus (\textbf{FrRCNN(ZF)})
\cite{Zeiler2014} and the deeper architecture of Simonyan and Zisserman 
(\textbf{FrRCNN(VGG16)}) \cite{Simonyan2015}. 
Throughout this work, we apply the \textbf{FrRCNN}
with default parameters learnt for the PASCAL VOC challenge. As we are
only interested in pedestrians we ignore all other classes and only pass
person detection results with output probabilities greater than 50\%
to the tracking framework. 

In our experiments, we found that the detection quality has a
significant impact on tracking performance when comparing the
\textbf{FrRCNN} detections to \textbf{ACF} detections. This is
demonstrated using a validation set of sequences applied to both an
existing online tracker \textbf{MDP} \cite{Xiang2015} and the tracker
proposed here. Table \ref{tbl:detector_compare} shows that the best
detector (\textbf{FrRCNN(VGG16)}) leads to the best tracking accuracy
for both \textbf{MDP} and the proposed method.

\begin{table}[tb!]
	\centering
	\caption{Comparison of tracking performance by switching the detector component. Evaluated on Validation sequences as listed in \cite{Xiang2015}.} 
	\scalebox{.8}
	{
	\begin{tabular}{llcccc}
		\toprule
		Tracker & Detector &\multicolumn{2}{c}{Detection} & \multicolumn{2}{c}{Tracking}\\
		& & Recall    & Precision & ID Sw & MOTA \\
		\midrule
		\multirow{3}{*}{MDP \cite{Xiang2015}} & ACF 	&	36.6 		& 	75.8		&	222	& 	24.0  		\\
		& FrRCNN(ZF)      & 46.2 & 67.2 & 	245	& 22.6	\\
		& FrRCNN(VGG16)      & \textbf{50.1} & 76.0 & 	\textbf{178}	& 33.5	\\
		\multirow{3}{*}{Proposed} & ACF    	&	33.6		& 	65.7		& 224	&  15.1	  		\\
		& FrRCNN(ZF)      & 41.3  & 72.4  & 347	& 24.0 \\
		& FrRCNN(VGG16)      & 49.5  & \textbf{77.5}  & 274	& \textbf{34.0} \\
		\bottomrule
	\end{tabular}
	}
	\label{tbl:detector_compare}
\end{table}

\subsection{Estimation Model}
Here we describe the object model, i.e. the representation and the
motion model used to propagate a target's identity into the next frame.
We approximate the inter-frame displacements of each object with a
linear constant velocity model which is independent of other objects and
camera motion.  The state of each target is modelled as:

$$
\mathbf{x} = [u,v,s,r,\dot{u},\dot{v},\dot{s}]^T,
$$
where $u$ and $v$ represent the horizontal and vertical pixel location
of the centre of the target, while the scale $s$ and $r$ represent the
scale (area) and the aspect ratio of the target's bounding box
respectively. Note that the aspect ratio is considered to be constant.
When a detection is associated to a target, the detected bounding box is
used to update the target state where the velocity components are solved
optimally via a Kalman filter framework \cite{Kalman1960}. If no detection is associated
to the target, its state is simply predicted without correction using
the linear velocity model.

\subsection{Data Association}

In assigning detections to existing targets, each target's bounding box geometry is estimated by predicting its new location in the current frame. The assignment cost matrix is then computed as the intersection-over-union (IOU) distance between each detection and all predicted bounding boxes from the existing targets. The assignment is solved optimally using the Hungarian algorithm. Additionally, a minimum IOU is imposed to reject assignments where the detection to target overlap is less than $IOU_{min}$.

We found that the IOU distance of the bounding boxes implicitly handles short term occlusion caused by passing targets.
Specifically, when a target is covered by an occluding object, only the occluder is detected, since the IOU distance appropriately favours detections with similar scale. 
This allows both the occluder target to be corrected with the detection while the covered target is unaffected as no assignment is made.

\subsection{Creation and Deletion of Track Identities}

When objects enter and leave the image, unique identities need
to be created or destroyed accordingly. For creating trackers, we
consider any detection with an overlap less than $IOU_{min}$ to signify
the existence of an untracked object. The tracker is initialised using
the geometry of the bounding box with the velocity set to zero. Since
the velocity is unobserved at this point the covariance of the velocity
component is initialised with large values, reflecting this uncertainty.
Additionally, the new tracker then undergoes a probationary period where
the target needs to be associated with detections to accumulate enough
evidence in order to prevent tracking of false positives.

Tracks are terminated if they are not detected for $T_{Lost}$ frames.
This prevents an unbounded growth in the number of trackers and
localisation errors caused by predictions over long durations without
corrections from the detector. In all experiments $T_{Lost}$ is set to 1
for two reasons. Firstly, the constant velocity model is a poor
predictor of the true dynamics and secondly we are primarily concerned
with frame-to-frame tracking where object re-identification is beyond
the scope of this work. Additionally, early deletion of lost targets
aids efficiency. Should an object reappear, tracking will implicitly
resume under a new identity.


\section{Experiments}
\label{sec:experiments}


We evaluate the performance of our tracking implementation on a diverse
set of testing sequences as set by the MOT benchmark database
\cite{MOT2015} which contains both moving and static camera sequences.
For tuning the initial Kalman filter covariances, $IOU_{min}$, and
$T_{Lost}$ parameters, we use the same training/validation split as
reported in \cite{Xiang2015}. The detection architecture used is the
\textbf{FrRCNN(VGG16)} \cite{Simonyan2015}.
Source code and sample detections from \cite{Simonyan2015} are available online.
\footnote{\url{https://github.com/abewley/sort}}

\begin{table*}[!tb] 
	\centering
	\caption
	{Performance of the proposed approach on MOT benchmark sequences \cite{MOT2015}.
		\vspace{1ex}  
	}
	\scalebox{1.0}
	{
		\begin{tabular}{l|cccccccccccc}
			\toprule
			 \textbf{Method} & \textbf{Type}    & \textbf{MOTA}$\uparrow$ & \textbf{MOTP}$\uparrow$  & \textbf{FAF}$\downarrow$ & \textbf{MT}$\uparrow$ & \textbf{ML}$\downarrow$ &  \textbf{FP}$\downarrow$  & \textbf{FN}$\downarrow$ & \textbf{ID sw}$\downarrow$  & \textbf{Frag}$\downarrow$      \\ 
			\midrule
			
			TBD \cite{Geiger2014PAMI} & Batch	& 15.9 & 70.9 & 2.6\%  &	6.4\% &	47.9\% &	14943 &	34777 &	1939 &	1963  	\\ 
			
			ALExTRAC \cite{Bewley2016_alextrac} & Batch	& 17.0 & 71.2 & 1.6\%  & 3.9\% & 52.4\% & 9233 & 39933 & 1859 & 1872   	\\ 
			
			DP\_NMS \cite{Pirsiavash2011} & Batch	&  14.5& 70.8& 2.3\%  & 6.0\% & 40.8\% &	13171&  34814&	 4537&	 3090&	  	\\ 
			
			SMOT \cite{Dicle2013} & Batch	& 18.2 &	71.2 & 1.5\%  & 2.8\% & 54.8\%	& 8780 &	40310 &	1148 &	2132  	\\ 
			
			NOMT \cite{Choi2015} & Batch	&  \textbf{33.7} & 71.9 &  \textbf{1.3\%}  & 12.2\% & 44.0\%	&  7762&	32547& \textbf{442} &	\textbf{823}  	\\ 
			
			RMOT \cite{Yoon2015} & \textbf{Online}	& 18.6 & 69.6& 2.2\%  & 5.3\% & 53.3\%	&  12473&	36835	 &	 684&	1282  	\\
			
			TC\_ODAL \cite{Bae2014}& \textbf{Online}	& 15.1 & 70.5 & 2.2\%  &	3.2\% &	55.8\% &	12970 &	38538 &	637 &	1716  	\\ 
			
			TDAM \cite{Min2015} & \textbf{Online}	& 33.0 & \textbf{72.8} & 1.7\%  & \textbf{13.3\%} & 39.1\%	& 10064 &	\textbf{30617} &	464 &	1506  	\\ 
			
			MDP \cite{Xiang2015} & \textbf{Online}	& 30.3 &	71.3 & 1.7\% & 13.0\% & 38.4\%	& 9717 & 32422 &	680 &	1500  	\\ 
			
			
			
			SORT (Proposed) & \textbf{Online}	& \textbf{33.4} & 72.1 & \textbf{1.3\%} & 11.7\% & \textbf{30.9\%} & \textbf{7318} & 32615 & 1001 & 1764   	\\ 
			\bottomrule 
			\noalign{\smallskip}
			
		\end{tabular}
		
	}
	\label{tbl:results}
	\vspace{0ex}
\end{table*}

\subsection{Metrics}

Since it is difficult to use one single score to evaluate multi-target tracking performance, we utilise the evaluation metrics defined in \cite{Li2009}, along with the standard MOT metrics \cite{Bernardin2008}:
\begin{itemize}[noitemsep,nolistsep]
	\item MOTA($\uparrow$): Multi-object tracking accuracy \cite{Bernardin2008}.
	\item MOTP($\uparrow$): Multi-object tracking precision \cite{Bernardin2008}.
	\item FAF($\downarrow$): number of false alarms per frame.
	\item MT($\uparrow$): number of mostly tracked trajectories. I.e. target has the same label for at least 80\% of its life span.
	\item ML($\downarrow$): number of mostly lost trajectories. i.e. target is not tracked for at least 20\% of its life span.
	\item FP($\downarrow$): number of false detections.
	\item FN($\downarrow$): number of missed detections.
	\item ID sw($\downarrow$): number of times an ID switches to a different previously tracked object \cite{Li2009}.
	\item Frag($\downarrow$): number of fragmentations where a track is interrupted by miss detection.
\end{itemize}
Evaluation measures with ($\uparrow$), higher scores denote better
performance; while for evaluation measures with ($\downarrow$), lower
scores denote better performance. True positives are considered to have
at least 50\% overlap with the corresponding ground truth bounding box.
Evaluation codes were downloaded from \cite{MOT2015}.

\subsection{Performance Evaluation}


Tracking performance is evaluated using the MOT benchmark \cite{MOT2015}
test server where the ground truth for 11 sequences is withheld. Table
\ref{tbl:results} compares the proposed method \textbf{SORT} with
several other baseline trackers. For brevity, only the most relevant
trackers, which are state-of-the-art online trackers in terms of
accuracy, such as (\textbf{TDAM} \cite{Min2015}, \textbf{MDP}
\cite{Xiang2015}), the fastest batch based tracker (\textbf{DP\_NMS}
\cite{Pirsiavash2011}), and all round \textit{near} online method
(\textbf{NOMT} \cite{Choi2015}) are listed. Additionally, methods which
inspired this approach (\textbf{TBD} \cite{Geiger2014PAMI},
\textbf{ALExTRAC} \cite{Bewley2016_alextrac}, and \textbf{SMOT} \cite{Dicle2013})
are also listed. Compared to these other methods, \textbf{SORT} achieves
the highest MOTA score for the online trackers and is comparable to the
state-of-the-art method \textbf{NOMT} which is significantly more
complex and uses frames in the near future. Additionally, as
\textbf{SORT} aims to focus on frame-to-frame associations the number of
lost targets (\textbf{ML}) is minimal despite having similar false
negatives to other trackers. Furthermore, since \textbf{SORT} focuses on
frame-to-frame associations to grow tracklets, it has the lowest number
of lost targets in comparison to the other
methods.


\subsection{Runtime}

Most MOT solutions aim to push performance towards greater accuracy,
often, at the cost of runtime performance. While slow runtime may be
tolerated in offline processing tasks, for robotics and autonomous
vehicles, realtime performance is essential. Fig.
\ref{fig:runtime_vs_mota} shows a number of trackers on the MOT
benchmark \cite{MOT2015} in relation to both their speed and accuracy.
This shows that methods which achieve the best accuracy also tend to be
the slowest (bottom right in Figure~\ref{fig:runtime_vs_mota}). On the
opposite end of the spectrum the fastest methods tend to have lower
accuracy (top left corner in Figure~\ref{fig:runtime_vs_mota}).
\textbf{SORT} combines the two desirable properties, speed and accuracy,
without the typical drawbacks (top right in
Figure~\ref{fig:runtime_vs_mota}). The tracking component runs at
\SI{260}{\Hz} on single core of an Intel i7 2.5GHz machine with 16 GB
memory. 


\section{Conclusion}
\label{sec:conclusion}

In this paper, a simple online tracking framework is presented that focuses on frame-to-frame prediction and association.
We showed that the tracking quality is highly dependent on detection
performance and by capitalising on recent developments in detection,
state-of-the-art tracking quality can be achieved with only classical
tracking methods. 
The presented framework achieves best in class performance with respect to both speed and accuracy, while other methods typically sacrifice one for the other.
The presented framework's simplicity makes it well suited as a baseline, allowing for new methods to focus on object
re-identification to handle long term occlusion. 
As our experiments highlight the importance of detection quality in tracking, future work will investigate a tightly coupled detection and tracking framework.

\vspace{1cm}

\addtolength{\textheight}{-1.1cm}  

\bibliographystyle{IEEEbib}

\end{document}